\def\year{2020}\relax
\documentclass[letterpaper]{article} 
\usepackage{aaai20}  
\usepackage{times}  
\usepackage{helvet} 
\usepackage{courier}  
\usepackage[hyphens]{url}  
\usepackage{graphicx} 
\usepackage{graphicx}  
\newcommand{\ev}{\mathbf{e}}

\newcommand{\relv}{\mathbf{rel}}

\newcommand{\betav      }{\boldsymbol \beta      }

\newcommand{\sigmav     }{\boldsymbol \sigma     }

\newcommand\bc{\textsc{BERT+ConvTransE} }
\newcommand\gc{\textsc{GCN+ConvTransE} }
\newcommand\sgc{\textsc{sim+GCN+ConvTransE} }

\newcommand\sgbc{\textsc{sim+GCN+BERT+ConvTransE} }

\title{Commonsense Knowledge Base Completion
with Structural and Semantic Context}
\author{Chaitanya Malaviya$^{\diamondsuit}$, 
Chandra Bhagavatula$^{\diamondsuit}$, 
Antoine Bosselut$^{\diamondsuit \clubsuit}$, 
Yejin Choi$^{\diamondsuit \clubsuit}$ \\
${}^{\diamondsuit}$Allen Institute for Artificial Intelligence\\
${}^{\clubsuit}$University of Washington\\
{\tt \{chaitanyam,chandrab\}@allenai.org}, {\tt \{antoineb,yejin\}@cs.washington.edu}
}
\newcommand{\citet}[1]{\citeauthor{#1} \shortcite{#1}}

\usepackage{todonotes}
\usepackage{amsmath}
\usepackage{amssymb}

\begin{document}

\maketitle

\begin{abstract}
Automatic KB completion for \emph{commonsense} knowledge graphs (e.g., ATOMIC and ConceptNet) poses unique challenges compared to the much studied conventional knowledge bases (e.g., Freebase). Commonsense knowledge graphs use free-form text to represent nodes, resulting in orders of magnitude more nodes compared to conventional KBs ( $\sim$18x more nodes in ATOMIC compared to Freebase (FB15K-237)). Importantly, this implies significantly sparser graph structures --- a major challenge for existing KB completion methods that assume densely connected graphs over a relatively smaller set of nodes.

In this paper, we present novel KB completion models that can address these challenges by exploiting the structural and semantic context of nodes. Specifically, we investigate two key ideas:
 (1) learning from local \emph{graph structure}, using graph convolutional networks and automatic graph densification
 and
 (2) \emph{transfer learning} from pre-trained language models to knowledge graphs for enhanced contextual representation of knowledge.
 We describe our method to incorporate information from both these sources in a joint model and provide the first empirical results for KB completion on ATOMIC and evaluation with ranking metrics on ConceptNet. Our results demonstrate the effectiveness of language model representations in boosting link prediction performance and the advantages of learning from local graph structure (+1.5 points in MRR for ConceptNet) when training on subgraphs for computational efficiency. Further analysis on model predictions shines light on the types of commonsense knowledge that language models capture well.\footnote{\hbox{Code and dataset are available at github.com/allenai/} commonsense-kg-completion.} 
\end{abstract}


\section{Introduction and Motivation}
\label{sec:intro}


\begin{figure}[t]
  \centering
  \includegraphics[width=0.95\columnwidth,height=6cm]{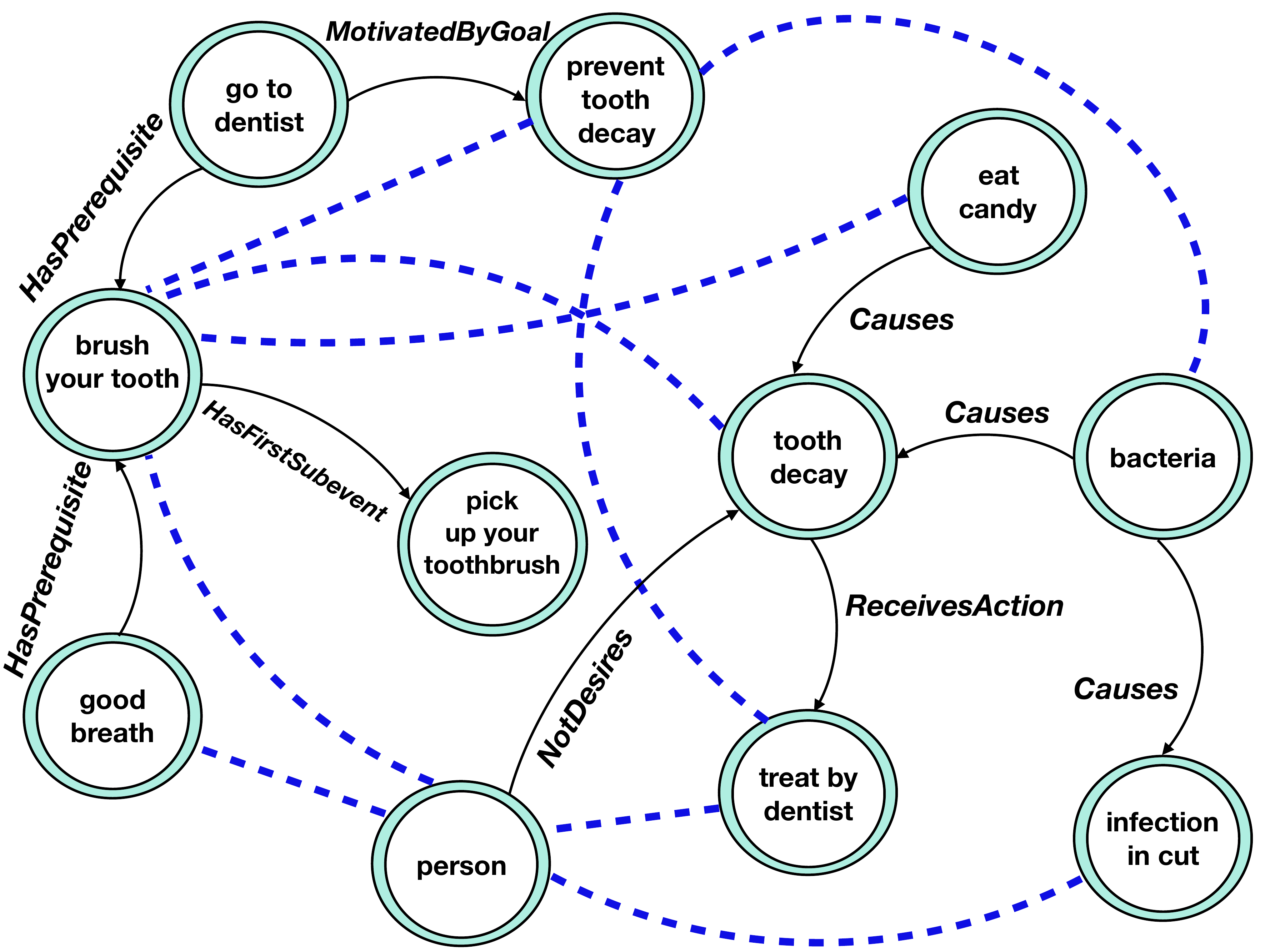}
  \caption{Subgraph from ConceptNet illustrating semantic diversity of nodes. Dashed blue lines represent potential edges to be added to the graph.}
  \label{fig:cn-graph}
\end{figure}

While there has been a substantial amount of work on KB completion for conventional knowledge bases such as Freebase, relatively little work exists for KB completion for \emph{commonsense} knowledge graphs such as ATOMIC \cite{sap2019atomic} and ConceptNet \cite{speer2013conceptnet}. The distinct goals of this paper are to identify unique challenges in commonsense KB completion, investigate effective methods to address these challenges, and provide comprehensive empirical insights and analysis. 

\begin{figure}
  \centering
  \includegraphics[width=0.9\columnwidth, height=3.2cm]{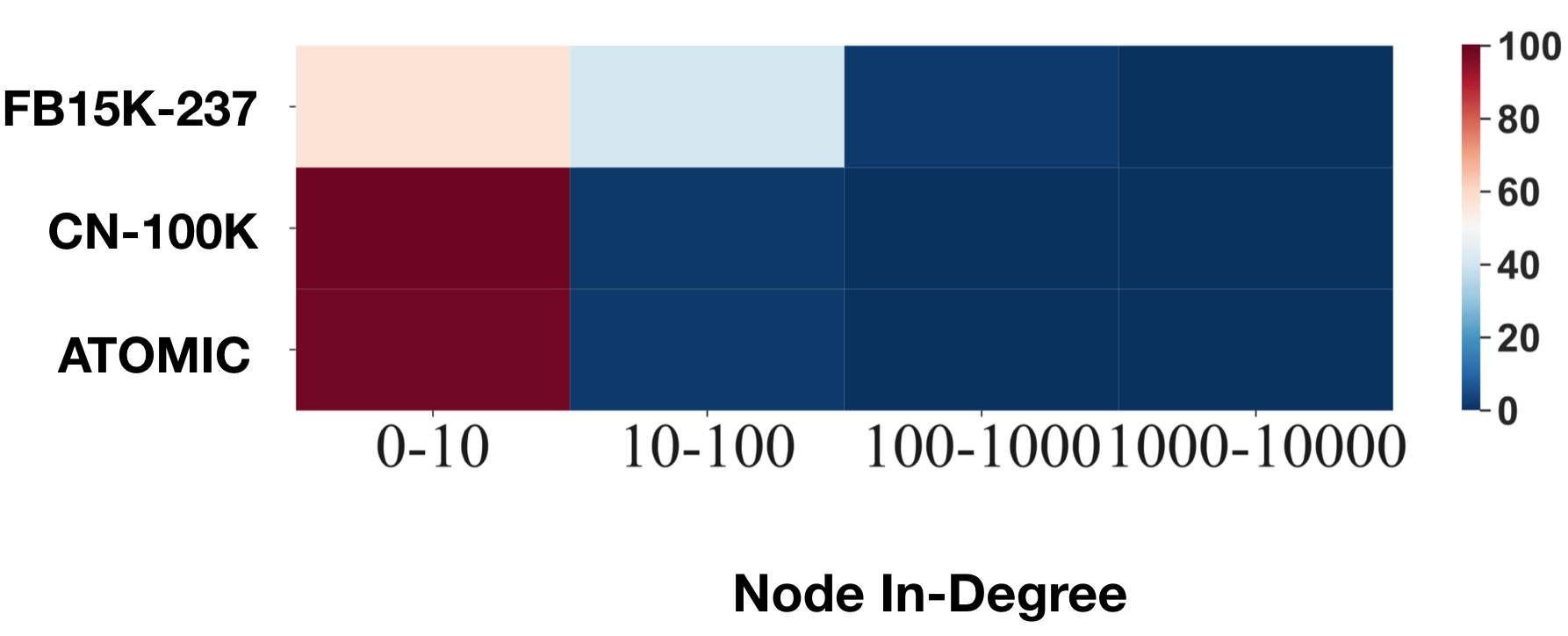}
  \caption{Heatmap showing the percentage of nodes with in-degree belonging to the specified bins on the x-axis. The plot illustrates the sparseness of commonsense KGs relative to a standard KB completion benchmark (FB15K-237).}
  \label{fig:degrees}
\end{figure}

The key challenge in completing commonsense KGs is the \emph{scale} and \emph{sparsity} of the graphs.
Unlike conventional KBs, commonsense KGs consist of nodes that are represented by \emph{non-canonicalized}, \emph{free-form} text, as shown in Figure~\ref{fig:cn-graph}. For example, the nodes "\textit{prevent tooth decay}" and "\textit{tooth decay}" are conceptually related, but \emph{not} equivalent, thus represented as distinct nodes. This conceptual diversity and expressiveness of the graph, imperative for representing commonsense, implies that the number of nodes is orders of magnitude larger, and graphs are substantially sparser than conventional KBs.
For instance, an encyclopedic KB like FB15K-237 \cite{toutanova2015observed} has 100x the density of ConceptNet and ATOMIC (node in-degrees visualized in Figure~\ref{fig:degrees}). 

\begin{figure}[t]
  \centering
  \includegraphics[width=\columnwidth,height=4.2cm]{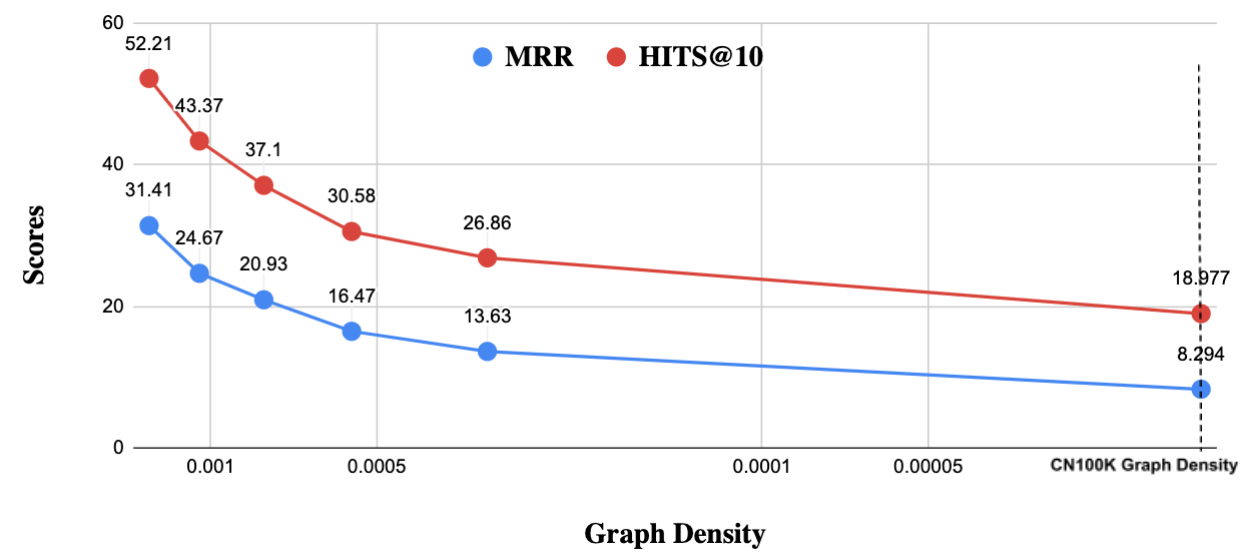}
  \caption{Trend of decreasing KB Completion Scores with different values of graph density (in log scale) for the FB15K-237 dataset with the ConvTransE model.}
  \label{fig:fb-graph}
\end{figure}

In this work, we provide empirical insights on how the sparsity of commonsense KGs poses a challenge to existing KB completion models that implicitly assume densely connected graphs. Figure~\ref{fig:fb-graph} provides a brief preview of this evidence, where the performance of ConvTransE \cite{shang2019end}, a high performing KB completion model, degrades quickly as we reduce the graph density of FB15K-237.

This motivates a strong need for investigating novel approaches to KB completion for commonsense KGs. 
We posit that new methods need to better accommodate the implicit conceptual connectivity across all nodes --- both \emph{structural} and \emph{semantic} -- beyond what is explicitly encoded in existing commonsense KBs. 
Specifically, we investigate two key ideas: 
 (1) learning from local \emph{graph structure}, using graph convolutional networks and automatic graph densification. 
 and 
 (2) \emph{transfer learning} from language models to knowledge graphs to improve \emph{contextual} representation of nodes.
 %

To integrate graph structure information, we present an approach based on graph convolutional networks (GCN) \cite{kipf2017semi} to contextualize a node's representation based on its local neighborhood.  
For transfer learning, we present effective approaches to fine-tune pre-trained language models \cite{devlin2019bert} to commonsense KGs, essentially achieving transfer learning from \emph{language} to \emph{knowledge}. 
Our work shares the high-level spirit of recent work from \citet{petroni19emnlp} that demonstrates the use of pre-trained LMs for reconstructing KB entries, but we provide a more focused study specifically for commonsense KGs.
%
Empirical analysis leads to the observation that GCNs, although effective on various densely connected graphs \cite{schlichtkrull2018modeling}, are not as effective on commonsense KGs out of the box, as sparse connections do not allow effective knowledge propagation. Hence, we propose an approach for automatic graph densification based on semantic similarity scores between nodes. Finally, we highlight strategies necessary to train models using information from both the graph structure and language models.


Our main contributions are highlighted below:
\begin{enumerate}
    \item Empirical insights about the unique challenges of commonsense KB completion compared to conventional encyclopedic KB completion. 
    \item Novel KB completion methods to model the implicit structural and semantic context of knowledge beyond what is explicitly available in existing KGs.
    \item The \emph{first} empirical results on ATOMIC for KB completion and evaluation with ranking metrics on ConceptNet.
    \item Analysis and insights on types of commonsense knowledge captured well by language models.
\end{enumerate}
In sum, our findings indicate that transfer learning is generally more effective than learning from graph structure. Moreoever, we find that graph structure can indeed provide complementary information which can boost performance, especially when training with subgraphs for efficiency.

\section{Knowledge Graphs}
There have been several efforts in building graph-structured representations of commonsense \cite{lenat1995cyc,speer2013conceptnet,cambria2014senticnet,sap2019atomic}. We focus our experiments on two prominent knowledge graphs: ConceptNet and ATOMIC. Statistics for both graphs are provided in Table~\ref{tab:stats}, along with FB15K-237 -- a standard KB completion dataset.


\begin{table*}[t]
\begin{center}
\begin{tabular}{c|ccccc}
\textbf{Dataset} & \textbf{\# Nodes} & \textbf{\# Edges} & \textbf{\# Relations} & \textbf{Density} & \textbf{Average In-Degree} \\  \hline
ConceptNet-100K & 78088   &   100000  &   34   &   1.6e-5  & 1.25 \\ 
\textsc{Atomic} & 256570 & 610536 & 9 & 9.0e-6 & 2.25 \\ 
FB15K-237 (for scale) & 14505 & 272115 & 237 & 1.2e-3 & 16.98 \\ \hline
\end{tabular}
\caption{\label{tab:stats} Knowledge Graph Statistics (Training Set Only). Graph density is calculated as $D={\frac  {V}{N (N-1)}}$, where $N$ is the number of nodes and $V$ is the number of edges in the graph. }
\end{center}
\end{table*}

\paragraph{ConceptNet-100K\footnote{\url{https://ttic.uchicago.edu/~kgimpel/commonsense.html}}: } CN-100K contains general commonsense facts about the world. This version \cite{li2016commonsense} contains the Open Mind Common Sense (OMCS) entries from ConceptNet \cite{speer2013conceptnet}.
The nodes in this graph contain 2.85 words on average. We used the original splits from the dataset, and combined the two provided development sets to create a larger development set. The development and test sets consisted of 1200 tuples each.
\paragraph{ATOMIC\footnote{\url{https://homes.cs.washington.edu/~msap/atomic/}}: } The ATOMIC knowledge graph contains social commonsense knowledge about day-to-day events. The dataset specifies effects, needs, intents and attributes of the actors in an event. The average phrase length of nodes (4.40 words) is slightly higher than that of CN-100K. Multiple targets may exist for a source entity and relation. The tuples in this graph may also contain a \texttt{none} target in the case that the relation type does not necessitate an annotation. 
The original dataset split was created to make the set of seed entities between the training and evaluation splits mutually exclusive. Since the KB completion task requires entities to be seen at least once, we create a new random 80-10-10 split for the dataset. The development and test sets consisted of 87K tuples each.



\section{Machine Commonsense Completion}

\begin{figure}[t]
  \centering
  \includegraphics[width=1\columnwidth]{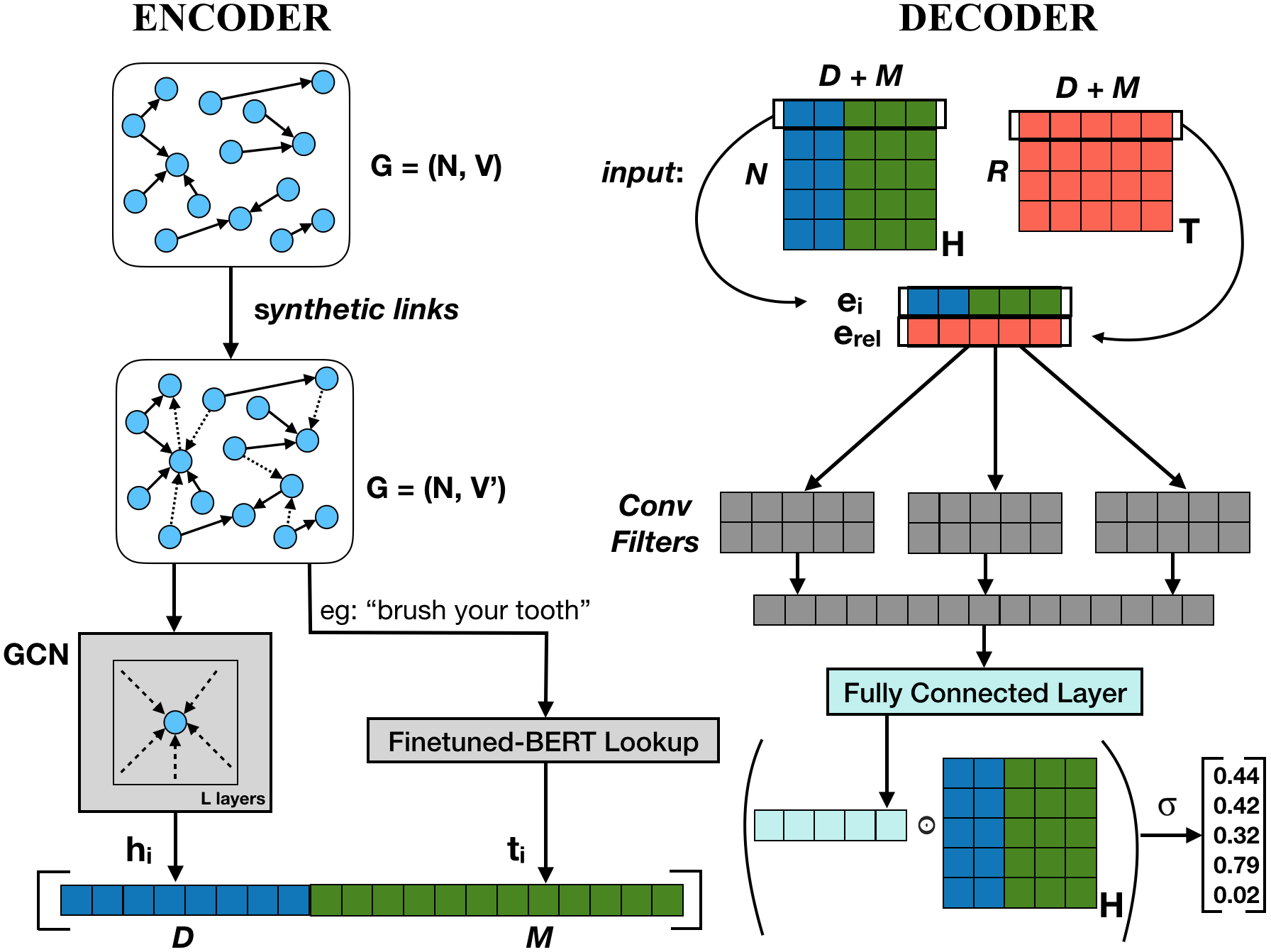}
  \caption{Model Architecture for machine commonsense completion.}
  \label{fig:model}
\end{figure}

We investigate two key ideas for performing completion of commonsense KGs -- 1) transfer learning from language to knowledge graphs and 2) learning from graph structure. To address the challenge of sparsity of commonsense KGs, we enrich the graph connectivity with synthetic semantic similarity links to enable the effective use of GCNs.  
The overall architecture of our model is illustrated in Figure~\ref{fig:model}.

\subsection{Problem Formulation}
Given a knowledge graph $G=(N,V)$ where $N$ is the set of nodes and $V$ is the set of edges, we consider a single training instance as the tuple $v_i = (\ev_1, \relv, \ev_2)$ with source entity $\ev_1$ represented in text as $\bar{\ev_1}$, relation type $\relv$ and target entity $\ev_2$, represented as $\bar{\ev_2}$.\footnote{we use the terms tuple and edge synonymously.} Here, $v_i \in V$ and $\ev_1,\ev_2 \in N$. The objective of the KB completion task is to maximize the score of a target entity $\ev_2$ given a tuple prefix $(\ev_1, \relv)$. Following previous work \cite{dettmers2018convolutional}, we also include inverse relations in our graph structure -- for every edge $(\ev_1, \relv, \ev_2)$, we add an inverse edge $(\ev_2, \relv^{-1}, \ev_1)$. 

\subsection{Transfer Learning from Text to Knowledge Graphs}
\label{ssec:model:phrase}
Transfer learning from language to knowledge graphs has recently been shown to be effective for commonsense knowledge graph construction \cite{bosselut19acl}. To transfer from language to knowledge graphs for completion, we finetune BERT \cite{devlin2019bert} with the masked language modeling loss and obtain rich semantic representations of nodes based on their text phrase.  This allows BERT to be attuned to a KG's specific style of text. The input for finetuning is the list of unique phrases used to represent nodes in the KG.  The format of the input to the model is \texttt{[CLS]} + $\bar{\ev_i}$ + \texttt{[SEP]}, where $\bar{\ev_i}$ is the natural language phrase represented by a node.
We use representations of the \texttt{[CLS]} token from the last layer of the BERT model as node representations in our models. We represent the node embedding matrix obtained from the BERT model as $T \in \mathbb{R}^{|N| \times M}$, where $M$ is the dimensionality of BERT embeddings.

\subsection{Learning from Graph Structure}
\label{ssec:model:gcn}
Graph Convolutional Networks (GCNs) \cite{kipf2017semi} are effective at incorporating information from the local neighborhood of a node in the graph.
A graph convolutional encoder take as input a graph $G$, and encodes each node as a $D$-dimensional embedding $h_i \in \mathbb{R}^D$ for all nodes $\ev_i \in N$.
The GCN encoder operates by sending messages from a node to its neighbors, optionally weighted by the relation type specified by the edge. This operation occurs in multiple layers, incorporating information multiple hops away from a node. The last layer's representation is used as the \textit{graph embedding} of the node. Several variants \cite{schlichtkrull2018modeling,velickovic2018graph} of these models have been proposed recently, all of which use the same underlying local neighborhood aggregation mechanism. We choose to use a version of the GCN which allows us to 1) parameterize the relation type corresponding to an edge and 2) account for the importance of a node's neighbor during aggregation. Given the graph $G$ with $R$ relation types and a GCN with $L$ layers, the operation for computing the node representation of a node $\ev_i$ in layer $l+1$ is: 

\begin{equation}
h_i^{l+1} = \tanh \left( \sum_{r \in R} \sum_{j \in J_i} \alpha_r \beta_{ij}^l W^l h_j^l + W_0^l h_i^l \right)
\label{eq:gcn}
\end{equation}

\noindent where $J_i$ represents the neighbors of the node $\ev_i$ in the graph, and $W^l$ is a linear projection matrix specific to layer $l$. The initial node representation $h_i^0$ is computed using an embedding layer. The second term in Equation~\ref{eq:gcn} represents the self-connection for the node, and is used to propagate information from one layer to the next. $\alpha_r$ is the weight for the relation type of the edge and $\beta_{i}^l$ is a vector denoting the relative importance of each of $\ev_i$'s neighbors:

\begin{equation}
\beta_{i}^l = \textnormal{softmax} (\hat{\betav_i^l})
\label{eq:softmax-sim}
\end{equation}
\noindent where each element of $\hat{\betav_i^l}$ is computed as,
\begin{equation}
\hat{\betav_{ij}^l} = h_i^l h_j^l
\label{eq:sim}
\end{equation}

\noindent Here, $h_i^l$ and $h_j^l$ are the representation of a node $\ev_i$ and its neighbor $\ev_j$. The output of the GCN is a node embedding matrix $H \in \mathbb{R}^{|N| \times D}$.

\subsubsection{Graph Densification}
\label{sec:syn}
The sparsity of commonsense KGs makes it challenging for GCNs to perform information propagation over a node's neighborhood. To tackle this issue, we add synthetic edges between nodes with semantically similar meanings to boost the learning of graph embeddings. These edges form a new synthetic \texttt{sim} relation and are only used for computing graph embeddings but not scored by the decoder. To form these edges, we use the fine-tuned BERT model described earlier to extract node representations and use these representations to compute the cosine similarity between all pairs of nodes in the graph. 

Upon computing the pairwise similarities, we use a hard threshold $\tau$ to filter the pairs of nodes that are most similar. This threshold is computed using different criteria for each graph, each prioritizing the precision of these links. For CN-100K ($\tau=$ 0.95 results in 122,618 \texttt{sim} edges), we plot the distribution of pairwise similarity values between all pairs of nodes and select the top $\sigmav / 2$ pairs of nodes to form these synthetic links, where $\sigmav$ is the standard deviation of the normally-distributed pairwise similarity distribution. For ATOMIC ($\tau=$ 0.98 results in 89,682 \texttt{sim} edges), we obtain a pairwise similarity distribution that is not normal, and hence use a threshold (measured up to 2 decimal points) that would only increase up to 100K edges in the graph\footnote{There are possibly other ways to choose the similarity threshold from a non-normal distribution, but we choose this criterion for the sake of simplicity. Decreasing this threshold resulted in minute drops in performance.}. After this step, we obtain a set of edges $V'$, where $|V'| > |V|$.

\subsection{Progressive Masking for Fusion} 
Models that use node embeddings from GCNs and BERT tend to overly rely on BERT embeddings, rendering graph embeddings ineffective
(we verify this using a random permutation test \cite{fisher2018model} where we randomly shuffle graph embeddings in a minibatch and observe little drop in performance). At the beginning of training, graph embeddings are not informative whereas fine-tuned BERT embeddings provide useful information -- causing the model to safely ignore graph embeddings. 
To prevent this issue, we randomly mask BERT embeddings starting with an all-zeros mask at the beginning to an all-ones mask at the end of 100 epochs\footnote{midway while training for 200 epochs.}. The ratio of dimensions masked is set as (epoch/100) for the first 100 epochs. This strategy forces the model to rely on both sources of information. A similar technique was used to enforce multimodal machine translation models to rely on images by masking out tokens in the source \cite{caglayan2019probing}.

\subsection{Convolutional Decoder}
\label{sec:convtranse}
Convolutional models provide strong scores for KB completion \cite{dettmers2018convolutional,shang2019end} and hence, we use a convolutional decoder. Given an edge prefix $(\ev_i, \relv)$, graph embeddings $H \in \mathbb{R}^{|N| \times D}$ from the GCN (\S\ref{ssec:model:gcn}) and BERT-based node embeddings $T \in \mathbb{R}^{|N| \times M}$ (\S\ref{ssec:model:phrase}), the decoder produces a score for $(\ev_i, \relv, \ev_j)$ where $\ev_i, \ev_j \in N$. We use the convolutional decoder \textsc{ConvTransE} \cite{shang2019end}, to score a tuple. This model is based on ConvE \cite{dettmers2018convolutional} but additionally models the translational property of TransE \cite{bordes2013translating}. 

The model uses one of the following as input node embeddings -- (i) graph embeddings $e_i = h_i$, (ii) BERT-based node embeddings $e_i = t_i$, or (iii) a concatenation of both $e_i = [h_i; t_i]$.\footnote{We experimented with other fusion methods (e.g., summation, linear transformation of concatenated representations, initializing GCN with BERT embeddings, tying initial graph embeddings with BERT embeddings), and found that concatenation as a fusion method works best.}
The model proceeds by first stacking the source node embeddings $e_i$, and relation embeddings $e_{rel}$, which are randomly initialized. The relation embeddings are chosen to have the same dimensionality as the embedding dimension of $e_i$, so that the stacking operation is possible. Assuming $C$ different kernels and $K$ as the width of a kernel, the output of kernel $c$ is given by,
\begin{align}
    m_c(\ev_i, \relv)[\eta] = \sum_{\tau=0}^{K-1} W_c (\tau, 0) e_i (\eta+\tau) & \nonumber\\
    + W_c (\tau, 1) e_{rel}(\eta+\tau)
\end{align}

We denote the output for the kernel $c$ to be $M_c \in \mathbb{R}^{|e_i|}$ and the concatenated output for all kernels to be $M \in \mathbb{R}^{C|e_i|}$. Finally, the score for a tuple $(\ev_i, \relv, \ev_j)$ is computed as,

\begin{equation}
s(\ev_i, \relv, \ev_j) = \sigma (M(e_i, e_{rel})W_{conv}e_j)
\label{eq:convtranse-score}
\end{equation}

where $W_{conv} \in \mathbb{R}^{C |e_i| \times |e_i|}$ is a bilinear projection matrix and $\sigma$ is the sigmoid function.
Upon computing scores for all candidate tuples $s(\ev_i, \relv, \ev_j)$, we use a binary cross entropy loss to train the model. All target nodes found in the training set are treated as positive instances while all non-existing target nodes are treated as negative instances. 

\subsection{Subgraph Sampling}
As the graph size increases, it becomes computationally intensive to train with the entire graph in memory. Specifically, it is intensive to perform graph convolution over the entire graph and compute scores for all nodes in the graph using the decoder. For instance, the model with GCN and BERT representations for ATOMIC occupies $\sim$30GB memory and takes 8-10 days for training on a Quadro RTX 8000 GPU.
Hence, we sample a smaller subgraph for training. We experiment with different sampling criteria and find that sampling edges uniformly at random provides the best performance.\footnote{We tried using random walk sampling, snowball sampling and uniform random sampling.} For graph densification, we link all pairs of nodes in the subgraph that cross the semantic similarity threshold $\tau$.




\section{Experimental Setup}
\label{sec:experiments}

\subsection{Evaluation Metrics}
Following previous work on KB completion \cite{yang2015embedding,dettmers2018convolutional}, we use ranking metrics (HITS and Mean Reciprocal Rank) for the purpose of evaluation.
Similar to \citet{dettmers2018convolutional}, we filter out all remaining entities valid for an $(\ev_1, \relv)$ pair (found across training + validation + test sets) from the ranking when computing scores for a gold target entity. The scores are measured in both directions, where we compute the ranking of $\ev_2$ given $(\ev_1, \relv)$ and the ranking of $\ev_1$ given $(\ev_2, \relv^-)$. We report the mean HITS and MRR scores averaged across both directions. 
We note that there are problems with these automatic metrics. Since commonsense KGs are highly sparse, several false negative target entities appear at the top of the ranking. Hence, we also perform human evaluation.
\subsection{Baselines}
We compare several high performing KB completion models and a transformer-based commonsense generation model. 

\paragraph{DistMult}
DistMult, proposed by \citet{yang2015embedding}, is an embedding-based method based on a bi-linear product between the entities and a relation matrix, which is constrained to be a diagonal matrix. The score function for a tuple is formulated as $s(\ev_1, \relv, \ev_2) = e_1^T W_{\relv} e_2$. 

\paragraph{ComplEx}
\citet{trouillon2016complex} proposed the use of complex-valued embeddings for nodes and relations, with the motivation that composition of complex embeddings can allow the model to handle a large number of relations. The model uses a tri-linear dot product as the scoring function.

\paragraph{ConvE}
ConvE \cite{dettmers2018convolutional} is a convolution-based model that stacks the node embedding and relation embedding for an entity prefix $(\ev_1, \relv)$ and reshapes the resulting tensor. The model performs 2D convolution upon this reshaped tensor and projects the output to a vector with the same dimensionality as the node embeddings.

\paragraph{ConvTransE}
\citet{shang2019end} proposed the \textsc{ConvTransE} model that builds upon the ConvE model but additionally models the translational properties of TransE. This model is discussed in depth in Section~\ref{sec:convtranse}.

\paragraph{COMeT}
\label{sec:comet}
We adapt a commonsense generation model \textsc{COMeT} \cite{bosselut19acl} for completion. \textsc{COMeT} generates the target phrase for a given (source entity, relation) prefix. 
Since \textsc{COMeT} was not trained using inverse relations, we only compute scores in the forward direction. We use the \textsc{COMeT} model by ranking target nodes based on the total and normalized negative log-likelihood score of each candidate tuple. In the absence of a standard approach for calibrating generation log-likelihoods, we report results using both metrics.\footnote{The total score can be biased against longer sequences; the normalized score can be biased against shorter sequences.}
The \textsc{COMeT} scores for ATOMIC have been computed on a smaller evaluation set with 2000 tuples due to computational limitations (denoted by * in table).


\subsection{Proposed Models}
All our proposed models use the \textsc{ConvTransE} decoder. We experiment with using a GCN as the encoder for obtaining node embeddings, these models are labeled with the prefix \textsc{GCN +}. Models that utilize synthetic links in the graph are labeled with the affix \textsc{sim +}. For models enriched with BERT representations, the \textsc{ConvTransE} decoder takes phrase representations extracted from BERT (\textsc{BERT +}) or a concatenation of graph embeddings and BERT embeddings (\textsc{GCN + BERT +}) as input node embeddings.

\subsection{Training Regimen}

We train all models for at least 200 epochs and continue training until the MRR on the development set stops improving. The MRR is evaluated on the dev set every 10 epochs for CN-100K and every 30 epochs for ATOMIC, and the model checkpoint with the highest MRR is used for testing. Further details about hyperparameter settings are specified in the supplemental material.


\begin{table*}[t!]
\begin{center}
\footnotesize
\begin{tabular}{@{\hskip 0in}l|c@{\hskip 0.05in}c@{\hskip 0.05in}c@{\hskip 0.1in}c|c@{\hskip 0.05in}c@{\hskip 0.05in}c@{\hskip 0.1in}c}
& \multicolumn{4}{c|}{\textbf{CN-100K}} & \multicolumn{4}{c}{\textbf{ATOMIC}} \\ 
& {\sc MRR} & {\sc Hits@1} & {\sc @3} & {\sc @10} & {\sc MRR} & {\sc Hits@1} & {\sc @3} & {\sc @10} \\
\hline
 \textsc{DistMult} & 
 8.97 & 4.51 & 9.76 & 17.44 &
 12.39 & 9.24 & 15.18 & 18.30 \\
 \textsc{ComplEx} & 
 11.40 & 7.42 & 12.45 & 19.01 &
 \textbf{14.24} & \textbf{13.27} & 14.13 & 15.96 \\
 \textsc{ConvE} & 
 20.88 & 13.97 & 22.91 & 34.02 &	
 10.07 & 8.24 & 10.29 & 13.37 \\
 \textsc{ConvTransE}  & 
 18.68 & 7.87 & 23.87 & 38.95 &	
 12.94 & 12.92 & 12.95 & 12.98 \\
 \textsc{COMeT-Normalized}  &
 6.07 & 0.08 & 2.92 & 21.17 &
 3.36* & 0.00* & 2.15* & 15.75* \\
 \textsc{COMeT-Total}  &
 6.21 & 0.00 & 0.00 & 24.00 &	
 4.91* & 0.00* & 2.40* & 21.60*  \\
 \hline
 \textsc{BERT + ConvTransE}  & 
 49.56 & 38.12 & 55.5 & 71.54 &	
 12.33 & 10.21 & 12.78 & 16.20 \\
 \textsc{GCN + ConvTransE}  & 
 29.80 & 21.25 & 33.04 & 47.50 &	
 13.12 & 10.70 & 13.74 & 17.68  \\
 \textsc{sim + GCN + ConvTransE} & 
 30.03 & 21.33 & 33.46 & 46.75 &
 13.88 & 11.50 & \textbf{14.44} & \textbf{18.38} \\
 \textsc{GCN + BERT + ConvTransE} & 
 50.38 & 38.79 & 56.46 & 72.96 &
 10.8 & 9.04 & 11.21 & 14.10 \\
 \textsc{sim + GCN + BERT + ConvTransE} & 
 \textbf{51.11} & \textbf{39.42} & \textbf{59.58} & \textbf{73.59} &
 10.33 & 8.41 & 10.79 & 13.86 \\
\end{tabular}
\end{center}
\caption{\label{tab:results-sub} KB Completion Results on CN-100K and ATOMIC with \textbf{subgraph sampling}. We present baselines in the top half of the graph and our implementations in the bottom half. '*' indicates difference in evaluation set described in Section~\ref{sec:comet}.}
\end{table*}

\begin{table}[ht]
\begin{center}
\footnotesize
\begin{tabular}{@{\hskip 0in}l|c@{\hskip 0.05in}c@{\hskip 0.05in}c@{\hskip 0.1in}c}
& \multicolumn{4}{c}{\textbf{CN-100K}}\\ 
& {\sc MRR} & {\sc Hits@1} & {\sc @3} & {\sc @10} \\
\hline
 \textsc{BERT + Conv..} & 
 52.25 & 41.04 & 58.46 & 73.50 \\
 \textsc{GCN + Conv..} & 
 27.24 & 18.96 & 30.46 & 43.17 \\	
 \textsc{sim + GCN + Conv..} & 
 27.51 & 19.04 & 30.79 & 45.46 \\
 \textsc{GCN + BERT + Conv..} & 
 50.80 & 39.29 & 57.33 & 72.66 \\
 \textsc{sim + GCN + BERT + Conv..} & 
 50.17 & 38.71 & 57.08 & 72.00 \\
\end{tabular}
\end{center}
\caption{\label{tab:results-full} KB Completion Results on CN-100K with \textbf{full graph training}. \textsc{Conv..} is ConvTransE.}
\end{table} 


\section{Results and Discussion}
\label{sec:res}

\subsection{KB Completion Performance}

We report our results on completion with subgraph sampling in Table~\ref{tab:results-sub}. For completeness, we also report results for CN-100K with the full graph in Table~\ref{tab:results-full}. The baseline models are trained with the full graph in memory.
We attempt to answer some pertinent research questions based on our results.

\subsection{What does BERT know about commonsense assertions?}

\textbf{Main Finding}: \textit{BERT is proficient at capturing taxonomic relations and hence, provides significant boosts for CN-100K but is not as effective for ATOMIC.}

When BERT representations are incorporated into a model, performance improvements ranging from $\sim$19-35 points on MRR are observed for CN-100K. This indicates that BERT can provide crucial information for the model to discriminate between target entities. We discuss the reasons for the usefulness of BERT representations below.

Assertions of the style present in CN-100K are prevalent in large text corpora, which indicates that masking of tokens during pre-training would have enabled BERT to already gain a headstart on the task of KB completion. For instance, consider a sentence "\textit{John bought plastic cups for the party.}" where we mask one token to obtain "\textit{John bought} [\texttt{MASK}] \textit{cups for the party.}" during pre-training. Now when the BERT representations are used to score a candidate tuple [\texttt{cups, MadeOf, plastic}], the model might compute a high score for this tuple because the BERT model solved a similar task during pre-training due to the masked LM objective. This result concurs with recent investigations on retrieval of commonsense knowledge from language models \cite{petroni19emnlp,feldman-davison-19emnlp}. Interestingly, BERT representations are found to be less useful for ATOMIC, because of a significantly larger number of nodes and more complex relation types (e.g. \texttt{oEffect} -- effect of event on others, \texttt{xNeed} --  what actor might need to do before event) in ATOMIC.

\begin{figure}[t]
  \centering
  \includegraphics[width=1\columnwidth, height=10cm]{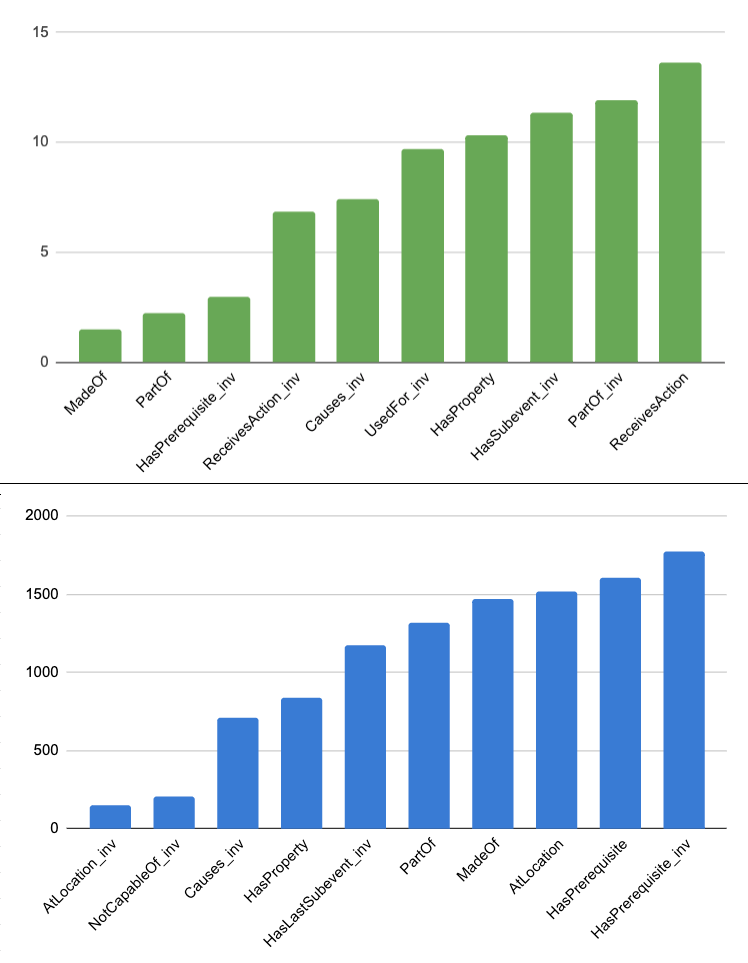}
  \caption{Top scoring relations for \textsc{BERT+ConvTransE} (top) and \textsc{sim+GCN+ConvTransE} (bottom).}
  \label{fig:rel}
\end{figure}

Finally, we attempt to use link prediction as a scaffold to find the commonsense relations that BERT captures well. Specifically, we use the \bc model trained for CN-100K to find the top scoring relations (by rank) in the evaluation set (presented in Supp:Figure~\ref{fig:rel}). We observe that BERT is most proficient at picking up on taxonomic relations such as \texttt{MadeOf}, \texttt{PartOf} and additionally, also does well on temporal relations such as \texttt{HasPrerequisite} and \texttt{ReceivesAction}. This can be explained by the higher frequency of taxonomic knowledge present in large text corpora as opposed to more complicated relations found in ATOMIC.

\subsection{How crucial is graph structure information?}


 \textbf{Main Finding}: \textit{Local graph structure information boosts performance when training with subgraph sampling, but improvements fade when training with the entire graph.}

We note that in the absence of BERT representations, incorporating graph embeddings provides strong improvements ($>$ +9 points on MRR for CN-100K and +$\sim$0.2 points for ATOMIC) over the model without a GCN encoder (i.e. \gc is better than \textsc{ConvTransE}). Similarly in the presence of BERT representations, the graph embeddings provide complementary information about the local neighborhood of a node which boosts performance for CN-100K. 
When training with the entire graph in memory, the improvements with using graph embeddings fade away. However, it is important to note that, using graph embeddings provides benefits when it is infeasible to train models with the entire graph in memory.

We further verify the importance of graph embeddings when BERT-based representations are used with a random permutation test \cite{fisher2018model}. In this test, we randomly shuffle the graph embeddings within each minibatch during inference and note the drop in performance. For the \sgbc model, we observe $\Delta$MRR drops of -7.24, -8.98 for CN-100K and ATOMIC, respectively.

\subsection{Do similarity-induced edges help?}


\textbf{Main Finding}: \textit{Similarity-induced edges boost learning of graph embeddings resulting in improved performance.}

We observe that when both graphs are trained with subgraph sampling but in the absence of BERT representations, augmenting the graph with similarity-induced edges provides improvements (\sgc $>$ \gc). When BERT representations are incorporated, the similarity-induced edges continue to help for CN-100K but not for ATOMIC. The \sgbc model achieves the best MRR for CN-100K. For ATOMIC, the \sgc model provides even stronger results than models which use BERT (+1.5 MRR points over \bc). This allows us to conclude that augmenting commonsense KGs with similarity-induced links can provide more context for computing graph embeddings.

\subsection{Can we use generative models for ranking tuples?}

\textbf{Main Finding}: \textit{Generative models cannot easily be repurposed to rank tuples for KB completion.}

Although generative models like \textsc{COMeT} have shown to produce diverse and precise target phrases, our results from Table~\ref{tab:results-sub} indicate that it is non-trivial to repurpose them to perform completion. This is partly due to the problems associated with using log-likelihood as an estimate for the truth of a tuple. Nonetheless, generative models such as \textsc{COMeT} have several merits. These models are faster to train, require lower memory for storage and are transductive in nature.
However, we argue that reasoning models that rely on KBs could benefit more from a discriminative approach towards commonsense knowledge induction, one that would make the graph denser without adding new nodes.

\subsection{Human Evaluation}
\label{sec:human}
Upon looking at model predictions, we note that false negative entities can appear at the top of the ranking. For instance, when $\ev_1$=\texttt{PersonX wins two tickets} and $\relv$=\texttt{xEffect} and the gold target entity $\ev_2$=\texttt{elated and excited}, our model predicts \texttt{elated} and \texttt{excited} separately in the top 10 candidate entities. Automated metrics fail to capture this semantic similarity in predictions. Hence, we perform human evaluation by presenting the top 10 predicted targets to 3 annotators on Amazon Mechanical Turk (AMT), for a subset of the instances in the dev set. This subset was created by sampling 200 instances randomly for CN-100K, and 450 instances (=50 instances for 9 relations) for ATOMIC. The annotators were asked to answer whether a tuple is valid, where a valid tuple is meaningful and true. Our results are reported in Figure~\ref{fig:human}. Samples of our model predictions with human annotations are provided in the supplementary material.
Human evaluation results indicate that using graph embeddings computed with graph densification in addition to BERT shows improvements. 

\begin{figure}[t]
  \centering
  \includegraphics[width=\columnwidth, height=3.5cm]{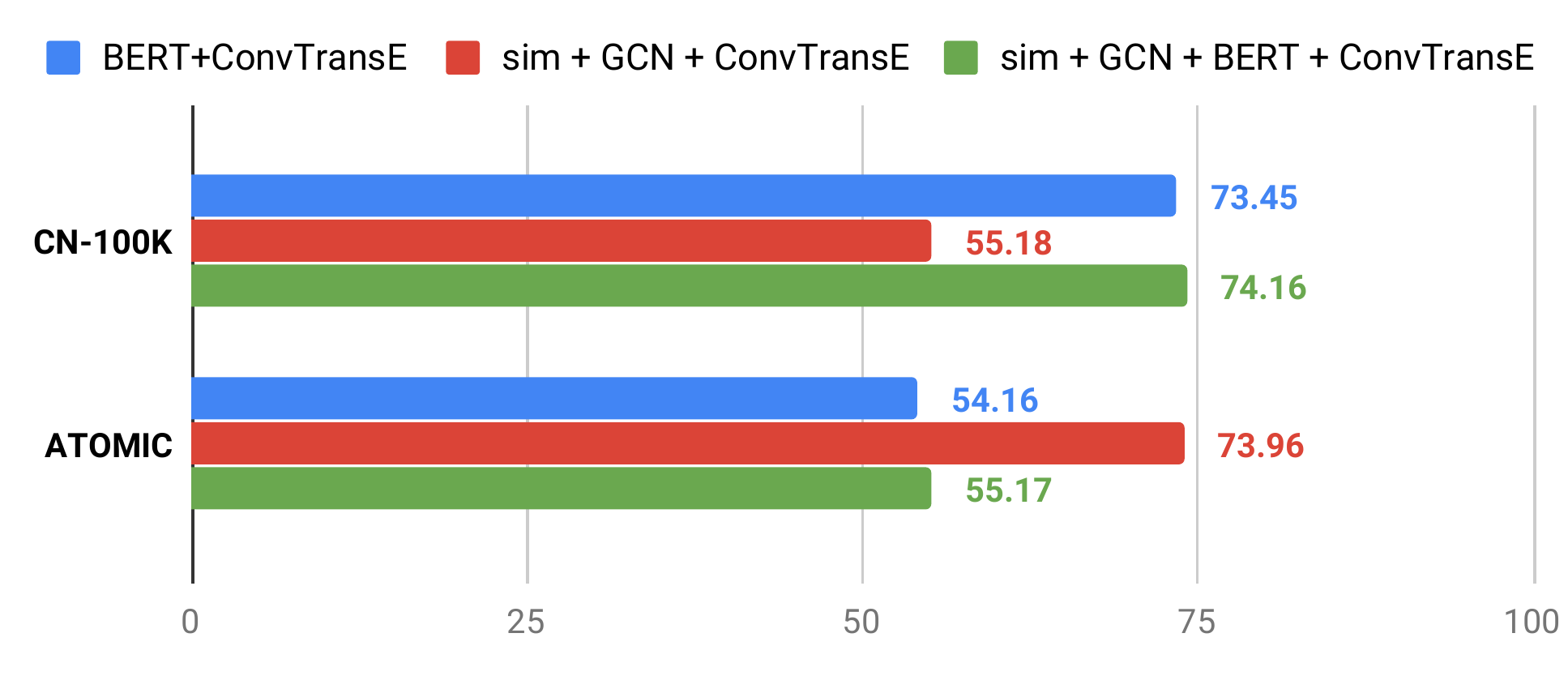}
  \caption{Human Evaluation Results (Average \% of \textit{valid} tuples among top-10 target candidates for random subset of entity-relation pairs)}
  \label{fig:human}
\end{figure}


\section{Related Work}
 A host of techniques \cite{bordes2013translating,yang2015embedding,trouillon2016complex,dettmers2018convolutional,shang2019end} have been proposed for KB completion. These can be classified into graph traversal based and embedding-based approaches. 
 Although embedding-based methods are more scalable, they usually assume enough training instances for each relation and a relatively high average degree for nodes to generalize well. Both these criteria are typically not satisfied by commonsense KGs.

 Among embedding-based methods, convolutional models have proven useful for computing entity-relation feature representations \cite{dettmers2018convolutional,shang2019end} and hence, we use a convolutional decoder. Simultaneously, we make use of GCNs to incorporate a node's neighborhood into its representation. Much as we do, \citet{shang2019end} use a GCN to compute graph embeddings and the ConvTransE decoder to score tuples. Our work differs from their model, as we account for the relative importance of neighbors in our graph convolution operation, tackle the sparsity of the graph structure by graph densification, and provide strategies to jointly train using graph embeddings and BERT representations. Finally, the use of BERT embeddings follows from work which showed that initializing node embeddings with pre-trained GloVe vectors \cite{glove} improves KB completion performance \cite{guu-etal-2015-traversing}.

Prior work on completing commonsense KBs \cite{li2016commonsense,saito2018commonsense,jastrzebski2018} uses BiLSTMs to encode a tuple and performs linear transformations to predict a score for binary classification. We argue that commonsense KB completion models should be trained and evaluated for ranking. A similar effort aimed at predicting new knowledge in ConceptNet, using dimensionality reduction was made by \citet{speer2008analogyspace}.


 We believe that completing commonsense KBs is bound to translate into improvements on a range of tasks that rely on these KBs, such as information retrieval \cite{kotov2012tapping}, question answering \cite{bauer2018commonsense,Musa2019AnsweringSE}, and reading comprehension \cite{ostermann2018semeval}. \citet{storks2019commonsense} provide a survey of the use of commonsense KBs in downstream tasks.


\section{Conclusion}
In this work, we show that existing KB completion models underperform with commonsense knowledge graphs, due to the sparsity and scale of these graphs. As our solution, we propose novel KB completion models, which are enriched by structural and semantic context, obtained from GCNs and language model representations. We describe a progressive masking strategy to efficiently utilize information from both sources. Further, we show that augmenting the graph with semantic similarity edges can help with completion. Our results indicate that 1) BERT-based node representations can provide significant improvements, especially when text in the graph is similar to pre-training corpora; 2) graph embeddings can provide rich local context for encoding nodes, and boost performance when training with subgraphs.




\section{Acknowledgments}
We thank the anonymous reviewers and the Mosaic team at AI2 for their insightful comments and suggestions. We also thank the Amazon MTurk workers who provided annotations to our human evaluation tasks. Finally, we thank the Beaker team at AI2 for providing support with experiments.
This research was supported in part by NSF (IIS-1524371, IIS-1714566), DARPA under the CwC program through the ARO (W911NF-15-1-0543), and DARPA under the MCS program through NIWC Pacific (N66001-19-2-4031). Computations on beaker.org were supported in part by credits from Google
Cloud.

\bibliographystyle{aaai}
\bibliography{aaai.bib}
\clearpage
\newpage


\section{Supplementary Material}


\subsection{Hyperparameter Details}
\label{sec:hyperparams}

\paragraph{BERT Fine-tuning} We used a maximum sequence length of 64, batch size of 32, and learning rate of 3e-5 to fine-tune the uncased BERT-Large model with the masked language modeling objective. The warmup proportion was set to 0.1.

\paragraph{Baseline Models} To train the baseline models, we used the implementations provided here.\footnote{https://github.com/TimDettmers/ConvE} We tuned the batch size and learning rate for the baseline models from {128, 256, 512} and {0.001, 0.002, 0.003} and used the default values for other hyperparameters.

\paragraph{Our Implementations} The graph convolutional network used 2 layers (using more (\{3,4,5\}) layers did not result in significant improvements) and an input and output embedding dimension of 200. The message passing algorithm for the GCN-based models was implemented using the Deep Graph Library (DGL). All embedding layers are initialized with Glorot initialization. The graph batch size used for subgraph sampling was 30000 edges. For the ConvTransE decoder, we used 500 channels, a kernel size of 5 and a batch size of 128. Dropout was enforced at the feature map layers, the input layer and after the fully connected layer in the decoder, with a value of 0.2. The Adam optimizer was used for optimization with a learning rate of 1e-4 and gradient clipping was performed with a max gradient norm value of 1.0. We performed L2 weight regularization with a weight of 0.1. We also used label smoothing with a value of 0.1. 


\subsection{Example Predictions}
\label{sec:samples}
\begin{figure}[ht]
  \centering
  \includegraphics[width=\columnwidth]{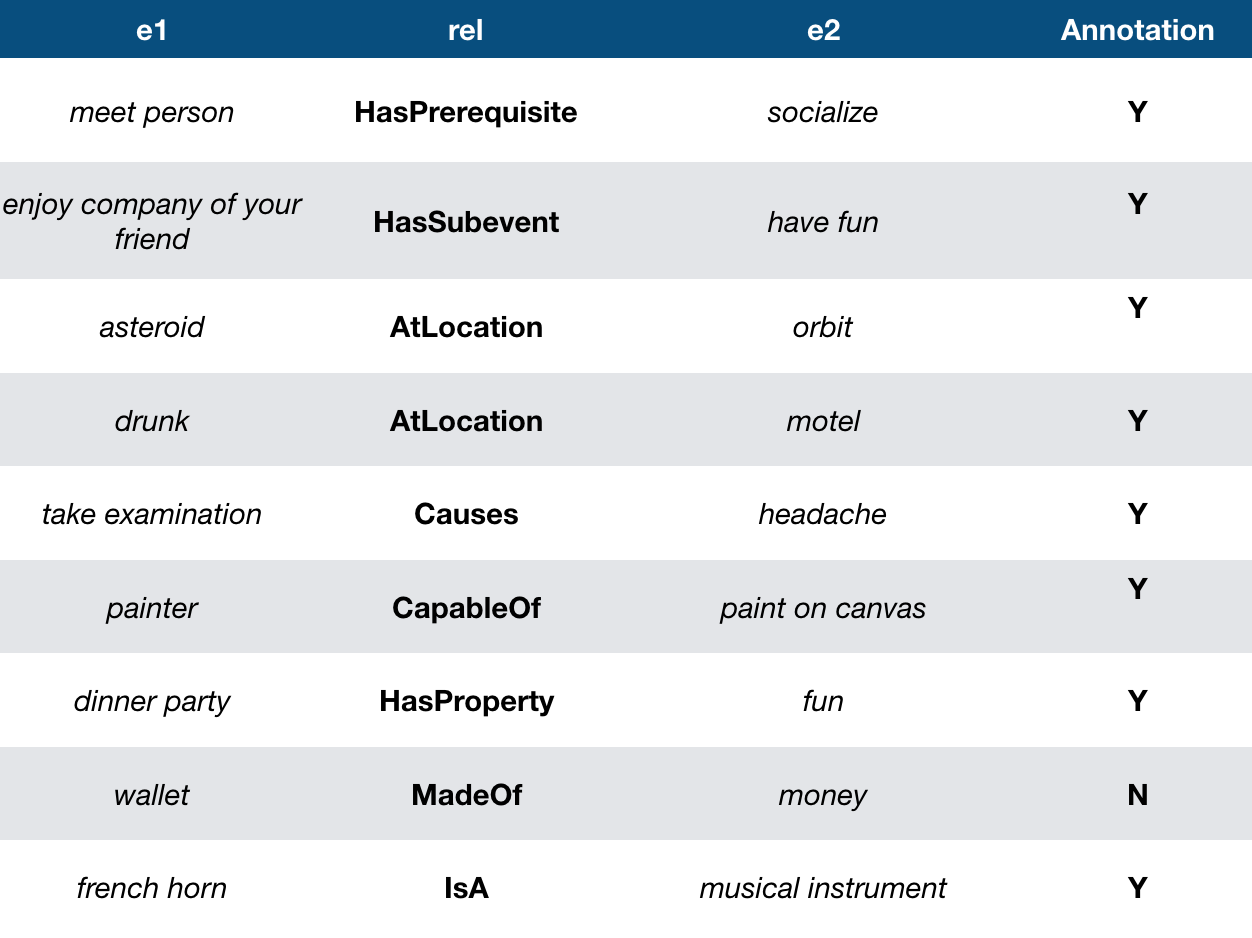}
  \caption{Randomly sampled top-1 predictions from best model (\textsc{sim+GCN+ConvTransE+BERT}) for CN-100K along with human annotations for validity of tuples.}
  \label{fig:outputs-cn}
\end{figure}

\begin{figure}[ht]
  \centering
  \includegraphics[width=\columnwidth]{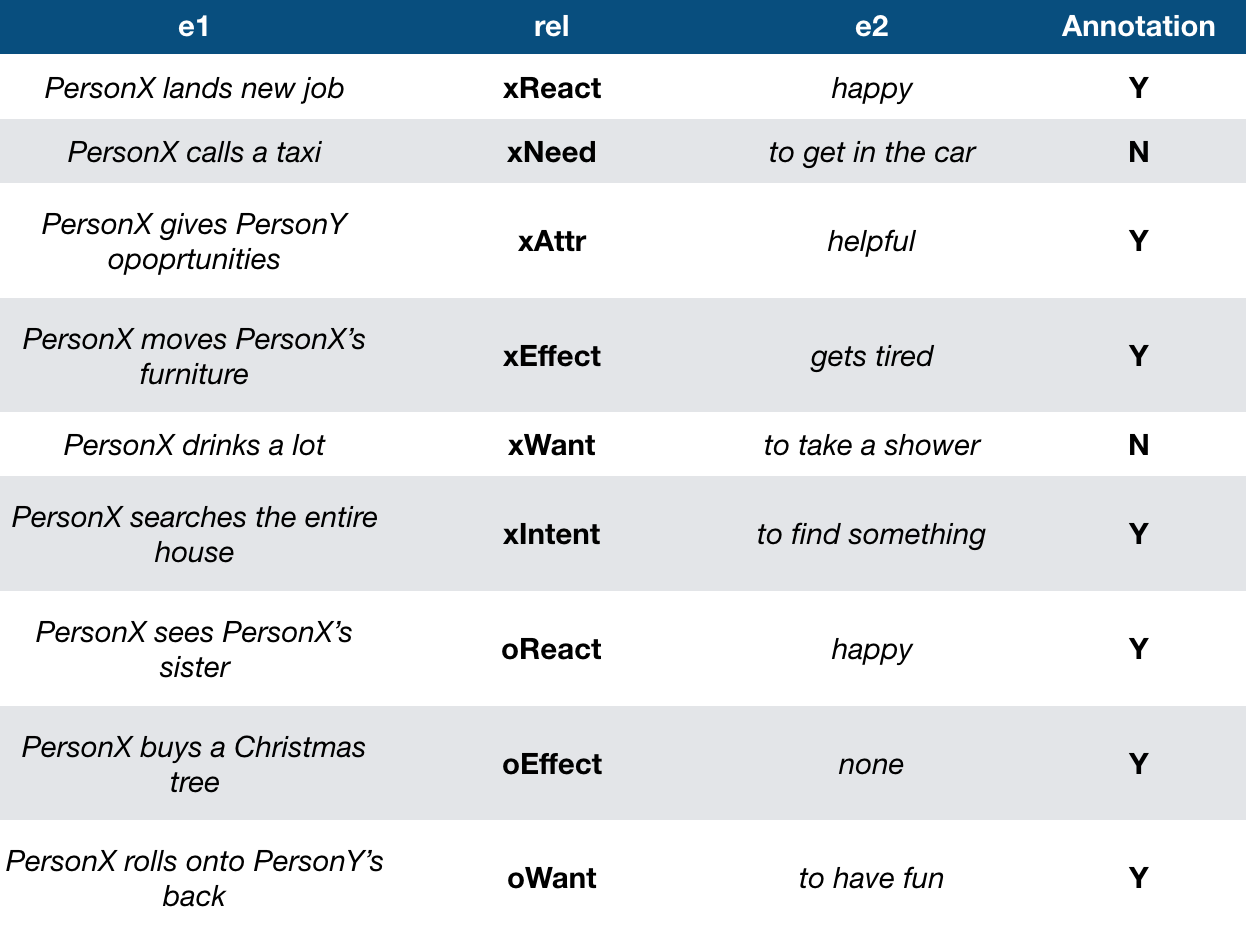}
  \caption{Randomly sampled top-1 predictions from best model (\textsc{sim+GCN+ConvTransE}) for ATOMIC along with human annotations for validity of tuples. While the complete relation descriptions can be found in \citet{sap2019atomic}, it is noted that "x" refers to PersonX and "o" refers to other actors besides PersonX.}
  \label{fig:outputs-atomic}
\end{figure}

\end{document}